\documentclass[10pt,twocolumn,letterpaper]{article}

\usepackage{wacv}              

\usepackage{graphicx}
\usepackage{amsmath}
\usepackage{amssymb}
\usepackage{booktabs}
\usepackage{import}
\usepackage{arydshln}
\usepackage{array}
\usepackage{footmisc}
\usepackage{multirow}
\usepackage[accsupp]{axessibility}  

\usepackage[pagebackref,breaklinks,colorlinks]{hyperref}
\usepackage[capitalize]{cleveref}
\crefname{section}{Sec.}{Secs.}
\Crefname{section}{Section}{Sections}
\Crefname{table}{Table}{Tables}
\crefname{table}{Tab.}{Tabs.}
\usepackage{caption}
\def\wacvPaperID{2312} 
\def\confName{WACV}
\def\confYear{2025}

\begin{document}

\title{Unified Framework for Open-World
Compositional Zero-shot Learning}

\author{Hirunima Jayasekara\\
\and
Khoi Pham\\
\and
Nirat Saini\\
\and
Abhinav Shrivastava\\
\and
University of Maryland\\
College Park\\
}
\maketitle

\begin{abstract}
Open-World Compositional Zero-Shot Learning (OW-CZSL) addresses the challenge of recognizing novel compositions of known primitives and entities. Even though prior works utilize language knowledge for recognition, such approaches exhibit limited interactions between language-image modalities. Our approach primarily focuses on enhancing the inter-modality interactions through fostering richer interactions between image and textual data. Additionally, we introduce a novel module aimed at alleviating the computational burden associated with exhaustive exploration of all possible compositions during the inference stage. While previous methods exclusively learn compositions jointly or independently, we introduce an advanced hybrid procedure that leverages both learning mechanisms to generate final predictions. Our proposed model, achieves state-of-the-art in OW-CZSL in three datasets, while surpassing Large Vision Language Models (LLVM) in two datasets. Our code is available at \url{https://github.com/hirunima/OWCZSL}


\end{abstract}

\newcommand{\et}{\textit{et al.}}
\newcommand\mypara[1]{\vspace{1mm}\noindent\textbf{#1}.}

\begin{figure*}[t]
  \centering
  \includegraphics[width=\textwidth]{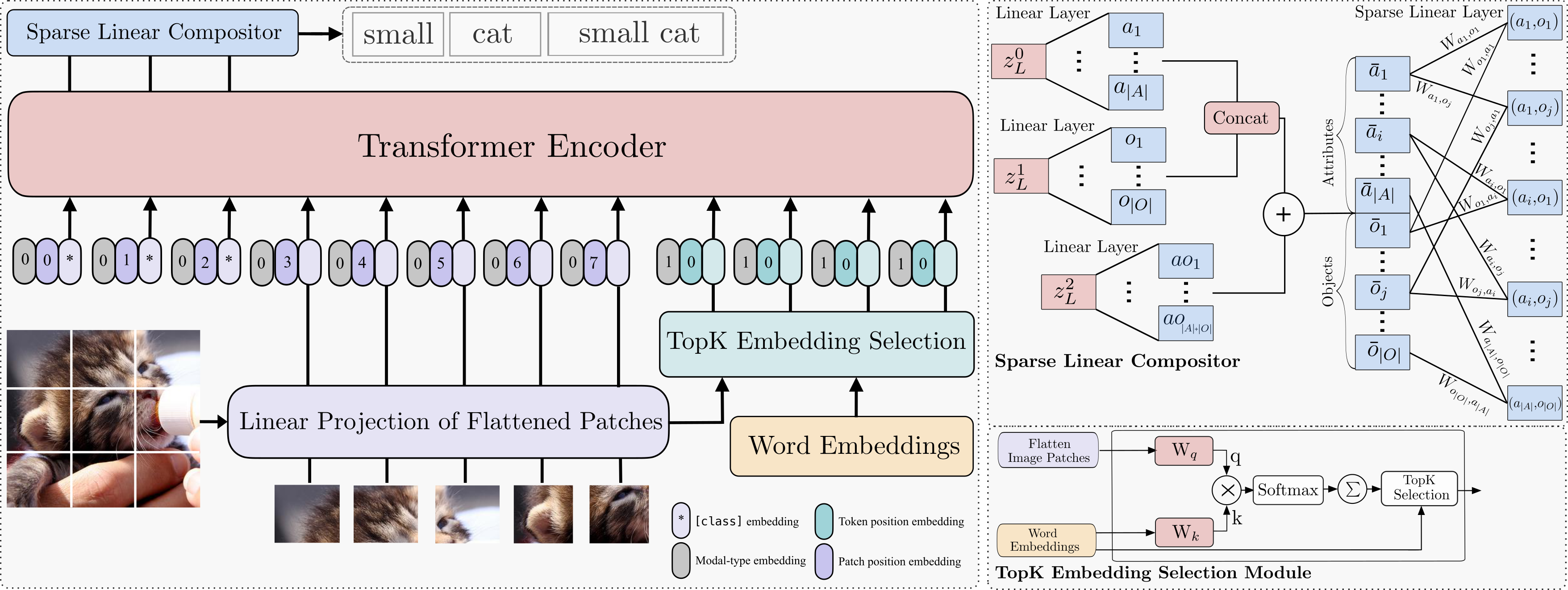}  
  \caption{The overall architecture of the proposed method. Input embeddings to the transformer encoder are formed by concatenating image patch embeddings and text embeddings. \textbf{TopK Embedding Selection Module} effectively selects relevant text embeddings that align with the provided image via cross attention. \textbf{Sparse Linear Compositor} computes attribute and object predictions alongside a final prediction vector utilizing a sparse linear layer.}
\label{fig:network}
\vspace{-0.6cm}
\end{figure*}

\section{Introduction}
\label{sec:introducton}
Historically, mankind has possessed cognitive capability to compose and decompose primitives and entities, thereby enabling generation of novel concepts as well as the interchangeability of primitives. For instance, a person may recognize a `Wet Cat` and use the attribute `Wet` to characterize a `Dog`. Additionally, the object `Cat` can encompass various attributes/descriptors such as `Black, Small or Sleepy`.  Compositional Zero-shot Learning (CZSL) introduces the idea of generating novel compositions by utilizing established primitives and entities.

Prior works can be categorized into two distinct groups based on the output prediction space, referred to as closed-world setting and open-world setting. Conventional CZSL methods utilize a closed-world setting \cite{Saini2022Disentangling,kim2023hierarchical,xu2021relation,wang2023learning,yang2020learning} in which, both seen and unseen pairs are present during inference evaluation; yet the output space only consists of a known number of seen and unseen pairs. Therefore, this setting requires prior knowledge of unseen pairs at the beginning of the inference, which hinders the capabilities of real time deployment. In order to mitigate this issue, Mancini \et \cite{mancini2021open} proposed a new setting where, for $|A|$ attributes and $|O|$ objects, the output space consists of all possible combinations of $|A|\cdot|O|$ pairs. This is coined as the open-world setting, and is considered a better evaluation criteria for real-world use cases.

In order to navigate the large label space in OW-CZSL, KG-SP \cite{karthik2022kg}, recognizes primitives (i.e., objects and attributes) separately. 
 Even though the lower cardinality of two sets reduce the number of computations required for composing pairs, the network does not necessarily learn pair compositions. In contrast, Nayak \et\cite{nayak2022learning} recognize compositions directly which may become challenging to handle as the number of attributes and objects increases. Thus, we try to combine the best capabilities of both paradigms through an advanced hybrid procedure wherein the proposed model learns primitives separately and compositionally using a novel Sparse Linear Compositor (SLC) module. 

Apart from visual features, the textual features assist in distinguishing primitives and entities respectively while incorporating semantic knowledge into the learning framework \cite{Saini2022Disentangling,kim2023hierarchical}. Therefore, the fusion of two modalities is more important when disentangling the compositions and generating new compositions. However, most notable works in CZSL \cite{mancini2021open,naeem2021learning,Saini2022Disentangling,karthik2022kg}, only facilitate simple fusion within two modalities. Their fusion approach typically involves projecting the embeddings of the two modalities into a shared embedding space and subsequently generating a similarity function between them. Furthermore, this requires the network to compute a similarity score for each potential composition during the inference. To address this, we introduce a TopK selection module which can automatically identify a subset of candidate text embeddings at the initial stage thereby reducing the computational overhead.

To that end, we hypothesize that a simple fusion of outputs even from high-performing unimodal embedders may prove inadequate to learn intricate vision-language based compositions \cite{kim2021vilt,nguyen2020movie,singh2022flava,bao2022vlmo}. This motivates our approach of finding an adequate inter-modal interaction method for compositional learning. Our proposed model seamlessly integrates dual modalities within a unified framework which reduces model overhead while maintaining high performance. Our proposed method is the first model in CZSL where both learnable embedding encoders are shallow allowing the model to focus the majority of computational resources on modeling interactions between modalities. 

Our key contributions are as follows:
\begin{itemize}
\setlength\itemsep{-0.1em}
\item We introduce a unified framework for OW-CZSL by utilizing a single transformer while preserving model performance to attain superior results compared to LLVMs showing state-of-the-art performance.
\item We propose TopK selection module to mitigate the challenge of exhaustively exploring all potential pairings during the inference process.
\item We propose a novel sparse linear layer to facilitate the integration of multiple tokens, aiding in the disentanglement of attributes and objects while enabling the generation of compositions with computational complexity reduced to $\mathcal{O}(|A|\cdot|O|)$.


\end{itemize}

\section{Related Works}
\label{sec:related_works}


\subsubsection{Compositional Zero-Shot Learning}
In contrast to Zero-Shot Learning (ZSL), Compositional Zero-Shot Learning (CZSL) necessitates the model to learn compositions of unseen concepts from already seen primitive components of the composition. The notion of compositional learning is introduced as a recognition problem \cite{hoffman1984parts}. However, Misra \et \cite{misra2017red} introduce the concept of composing objects and attributes by employing standalone classifiers for the primitive components of a given object which are then integrated into a final classifier.

In order to facilitate generalization into unseen pairs, a common approach in prior works involves employing a shared embedding space for both visual and textual features. This enables the composition and decomposition of primitives while maintaining flexibility in the model's output. This enables a wide array of research focus on exploring various methods to extract dual embedding features where, utilizing attributes as operates to modify objects \cite{nagarajan2018attributes}, hierarchical decomposition-and-composition of attributes \cite{yang2020learning}, encoding conditional attributes based on recognized object \cite{wang2023learning}, Blocked Message Passing Network to capture the relations between primitive concepts \cite{xu2021relation} and CoT \cite{kim2023hierarchical} deploying visual network hierarchy to generate representative embeddings.
\vspace{-0.3cm}
\subsubsection{Open World Compositional Zero Shot Learning}
\vspace{-0.1cm}
A significant challenge within OW-CZSL pertains to effective management of the output space. In addressing this, KG-SP \cite{karthik2022kg} proposes two separate image encoders for attributes and objects, while incorporating visual and textual embeddings by projecting to a common embedding space. SAD-SP \cite{li2023distilled} introduces attention between contextual features to the OW-CZSL setting in order to disentangle objects and attributes followed by DRANet \cite{liu2023simple} which utilizes attention in visual cues, adopting non-local attention to capture contextuality and using local attention to enhance visual distinction. Moreover, SymNet \cite{li2020symmetry} uses symmetry learning for coupling and decoupling of attributes. Nayak \et \cite{nayak2022learning} integrates CLIP \cite{radford2021learning} into the context of OW-CZSL by employing soft-prompting \cite{zhou2022learning} to recognize object attribute pairs. 
Although CLIP demonstrates impressive zero-shot performance, the suitability of adopting it as a foundational model for (OW-CZSL) is subject to debate, primarily due to the possibility that CLIP might have encountered unseen pairs during pre-training. Therefore, we refrain from incorporating CLIP into our framework. ADE \cite{hao2023learning} successfully implements a ViT \cite{dosovitskiy2020image} based image feature extractor for OW-CZSL; however, ADE uses a similar dual stream architecture as previous methods in the literature. To the best of the authors’ knowledge this is the first attempt on deploying a ViT based unified network for OW-CZSL.
Similar to prior works on open world \cite{karthik2022kg,liu2023simple}, we predict objects and attributes separately. However, while their final pair predictions are composed in isolation, we manually entangle and disentangle attributes and objects with assistance of textual knowledge to achieve better generalization.
\vspace{-0.4cm}
\subsubsection{Pair Feasibility}
\vspace{-0.1cm}
 Given the considerably large output space, it is advantageous to eliminate unfeasible attribute-object pairs (e.g., ripe cat). By leveraging external knowledge to measure the compatibility between attributes and objects, a feasibility score can be calculated for each pair. A naive approach would be computing n-gram language models \cite{brown1992class} for each pair using a large-generalizable corpus. Another approach would entail computing the similarity measures between the attribute and the object using word embeddings \cite{nayak2022learning,karthik2022kg} that contain commonsense knowledge such as GloVe \cite{pennington2014glove} or ConceptNet \cite{speer2017conceptnet}. We adopt the latter approach by deploying a combine feasibility score computation schema which integrates knowledge from both GloVe and ConceptNet word embeddings.
\vspace{-0.4cm}
\subsubsection{Multi Modality with Transformers}
\vspace{-0.1cm}
Multi modality based transformers are quite popular among several vision areas: such as object detection \cite{kim2021vilt,li2022grounded}, recognition \cite{radford2021learning,singh2022flava,yang2022unified}, attribute prediction \cite{pham2022improving}, Visual Question Answering \cite{kim2021vilt,singh2022flava}, image captioning \cite{yu2022coca}, and image generation\cite{ramesh2021zero}. The intertwinement of two modalities can facilitate with inter-modal or intra-modal interactions across layers. Aforementioned interaction schema can split into two categories \cite{bugliarello2021multimodal}: 1) single-stream approaches where input consists of concatenation of image and text inputs to enable inter-modal interactions throughout all the layers; (e.g., VisualBERT \cite{li2019visualbert}, UNITER \cite{chen2019uniter}, ViLT \cite{kim2021vilt})  and 2) dual-stream approaches  where the two modalities encode separately and interact at later layers.(e.g., ViLBERT \cite{lu2019vilbert}, LXMERT \cite{tan2019lxmert}, and CLIP \cite{radford2021learning}). CLIP uses two separate transformer embedders for each modality. Prediction calculation between the pooled image vector and text vector is computed by dot product which provide minimal inter-modal interactions. In our proposed work, we explore a single-stream approach aimed at facilitating a maximum amount of inter-modal interactions.

\section{Method}
\label{sec:method}

\subsection{Problem Formulation}
We follow the open world setting proposed by Mancini \et \cite{mancini2021open} for OW-CZSL. Each training sample consists of two elements, Image $x \in X$ and corresponding text label $t=(t_{attr},t_{obj}) \in T$, where $t_{attr}$ and $t_{obj}$ are text labels of attribute and object respectively followed by attribute and object labels, $y=(y_{attr},y_{obj}) \in Y$. $Y$ consists of two subsets, seen pairs $Y^s$ which, contains all attribute and object compositions present during training and unseen pairs $Y^u$ which consists of all attribute and object compositions that were not available during training resulting in $Y^s \cup Y^u=Y $ and $Y^s \cap Y^u=\emptyset$. Similar to Saini \et \cite{Saini2022Disentangling} each image is followed by two other images sampled from training data, in which the first image $x_{obj}$ includes the same object as $x$ and a different attribute and the second image $x_{attr}$ includes the same attribute as $x$ and a different object. Lastly, for OW-CZSL setting, we set our label space to be every possible attribute ($A$) and object ($O$) composition ($|A|\cdot|O|$).

\subsection{Method Overview}
Our goal is to develop a simple and scalable OW-CZSL model. Therefore, we
focus on standard Transformer-based models due to their scalability \cite{dosovitskiy2020image} and simplicity compared to dual transformer models such as CLIP \cite{radford2021learning}. Our proposed method explores the extent to which the capacity of the inter modality interactions contribute to the overall performance of the OW-CZSL. As illustrated in Figure \ref{fig:network}, our model employs a standard Vision Transformer (ViT)\cite{dosovitskiy2020image} as the dual modality encoder.

\subsection{Visual and Language Embeddings}
The initial stage involves generating separate embeddings for visual and language features, subsequently concatenating and feeding the aggregated feature representation into the transformer encoder.
\vspace{-0.4cm}
\subsubsection{Linear Patch Projection}
\vspace{-0.1cm}
We utilize linear flattened patch projection schema from ViT \cite{dosovitskiy2020image} where, the image $x \in \mathbb{R}^{H\times W\times C}$ and flattened into to $v \in \mathbb{R}^{N\times(P^2\cdot C)}$ with $N=HW/P^2$ number of $P \times P$ patch projections followed by linear projection $V \in \mathbb{R}^{(P^2\cdot C)\times H}$ of $v$, producing $x_v \in \mathbb{R}^{N\times(P^2\cdot C)}$ as visual embeddings for the network. Even though it is not standard practice to learn attribute and object predictions as separate entities, we follow a similar thought as Misra \et \cite{misra2017red} and consider OW-CZSL as a multi task learning problem \cite{misra2016cross} resulting in three $\mathtt{[class]}$ tokens for attributes, objects and pairs. Lastly, positional embeddings $V^{pos}\in \mathbb{R}^{(N+1)\times H}$ are embedded to the patch embeddings.
\begin{align}
 x_v& = [v_1V ; \dotsm ; v_NV ]\\
\tilde{v}& = [v_{attr}; v_{obj}; v_{pair}; v_1V ; \dotsm ; v_NV ] + V^{pos}
\end{align}

\subsubsection{Language Embeddings} We create a fixed auxiliary vocabulary input for the dataset by collecting BERT embeddings \cite{kenton2019bert} for each attribute or object in $T$. We opt for BERT as the text encoder due to its ability to provide contextualized embeddings as BERT's enriched semantic understanding surpasses that of uni-modal text embedders \cite{chen2023difference}. For each text, the last embedding output $u^{attr}_i \in \mathbb{R}^{(|A|)\times d}, u^{obj}_i \in \mathbb{R}^{(|O|)\times d}$ is extracted and concatenated to form a vocabulary embedding matrix $u_{vocab}\in \mathbb{R}^{(|A| + |O|)\times d}$ where d is the dimension of the vocabulary embedding and $d=P^2\cdot C$.
\begin{align}
u_{vocab}& = [u^{attr}_1; \dotsm ;u^{attr}_{|A|};u^{obj}_1;  \dotsm ; u^{obj}_{|O|} ]
\end{align}

\subsection{TopK embedding Selection}
The objective of this module is to perform visual-assisted vocabulary mapping to create text input for the primary multi-modality transformer. 
To represent a class, we combine two text embeddings extracted from the vocabulary: one pertaining to attribute and the other to object. 

For brevity, we explain TopK attribute selection while TopK object selection follows the same procedure. 
In order to select the most descriptive and relevant attribute lexicon corresponding to the given image, we use respective image embeddings $\tilde{v}$ as the Query vector and attribute portion of text embeddings $u_{vocab}[0:|A|]$ as the Key vector. We impose cross attention between Query and Key to produce an attention map, $\mathcal{A}_a  \in \mathbb{R}^{(P^2\cdot C) \times |A|}$ .
\begin{align}
\mathcal{A}_a &= \text{softmax}\left(\frac{\tilde{v}\cdot u_{vocab}[0:|A|]}{\sqrt{P^2\cdot C}}\right)
\end{align}
As illustrated in Figure \ref{fig:network}, the final score is calculated by summing $\mathcal{A}_a$ along the image axis to produce an 1-D vector consisting of attention scores, $\tilde{\mathcal{A}}_a \in \mathbb{R}^{|A|}$ for each word. 
\vspace{-10pt}
\begin{align}
\tilde{\mathcal{A}}_a &= \sum_{i=0}^{P^2·C}{\mathcal{A}_a[i]}
\end {align}
From sorted $\tilde{\mathcal{A}}$, we select text embeddings of words which has top K highest attention scores. The same procedure is repeated for selecting top K object embeddings by utilizing the sorted attention scores corresponding to objects $\tilde{\mathcal{A}}_o \in \mathbb{R}^{|O|}$. Text input $\tilde{u}\in \mathbb{R}^{2K}$ to the multi modality transformer is created by 
\begin{align}
\tilde{u}& = [u^{attr}_{\tilde{\mathcal{A}}_a[1]};\dotsm ;u^{attr}_{\tilde{\mathcal{A}}_a[K]}; u^{obj}_{\tilde{\mathcal{A}}_o[1]};\dotsm ; u^{obj}_{\tilde{\mathcal{A}}_o[K]} ]
\end{align}
\subsection{Language based Visual Modality Transformer}
Similar to Kim \et\cite{kim2021vilt}, we initialize the transformer weights from pre-trained ViT\cite{dosovitskiy2020image} weights rather than that of BERT. This expects to bolster the model's ability to process visual features effectively, thereby addressing the challenge of lacking a separate uni-model visual embedder.

In order to separate two modalities, text and image embeddings are combined with their respective modal-type learnable embedding vectors, denoted as $v^{type}$ and $u^{type}$, where $ u_{type}, v_{type} \in \mathbb{R}^{P^2\cdot C} $ then concatenated along the embedding axis to form input sequence $z^0\in \mathbb{R}^{M\times P^2\cdot C}$ where, $M=N+2K+3$ is total number of input tokens.
\vspace{-10pt}
\begin{align}
z^0 &= [\tilde{v}+ v^{type}; \tilde{u}+ u^{type}]
\end{align}
Following suit of Dosovitskiy \et\cite{dosovitskiy2020image}, contextualized vector $z^0$ is iteratively updated through D number of transformer layers with multi-head self-attention (MSA) to obtain the final contextualized vector $z^d \in \mathbb{R}^{M\times P^2\cdot C}$. 
\begin{align}
z^d &= MSA(MLP(z^{d-1}))+z^{d-1},\hspace{0.6cm} d=1\dotsm D
\end{align}
For all experiments, we use weights from ViT-B/16 pre-trained on ImageNet-21K\footnote{We expect by utilizing weights that are pre-trained on larger datasets (e.g., WIT, JFT-300M) would improve the performance of the model.}, with hidden size $P^2\cdot C$ of 768, layer depth D is 12, patch size P is 16, and the number of attention heads is 12.

\vspace{-2mm}
\subsubsection{Sparse Linear Compositor}
\vspace{-1mm}
Given three $\mathtt{[class]}$ tokens, Sparse Linear Compositor computes an attribute prediction vector, an object prediction vector as auxiliary outputs and a composition vector between $|A|$ the number of attributes and $|O|$ the number of objects. To derive auxiliary outputs, the first two $\mathtt{[class]}$ tokens are processed through two learnable MLP heads. 
\begin{align}
\tilde{y}_{\text{attr}} &= MLP_{\text{attr}}(z_0^D) \quad\text{and} \quad\tilde{y}_{\text{obj}} = MLP_{\text{obj}}(z_1^D)
\end{align}
Where, $\tilde{y}_{\text{attr}} \in \mathbb{R}^{|A|}$ and $\tilde{y}_{\text{obj}} \in \mathbb{R}^{|O|}$. In order to compute the decomposition pair predictions, we normalize the linear head outputs and multiply to create $\tilde{y}_{\text{decompose}} \in \mathbb{R}^{(|A|\cdot |O|)}$
\begin{align}
\tilde{y}_{\text{decompose}} &= Norm(\tilde{y}_{\text{attr}}) \times Norm(\tilde{y}_{\text{obj}})
\end{align}
Third $\mathtt{[class]}$ token is combined with the row wise concatenation of attribute and object predictions to produce $\tilde{z}_{\text{pair}} \in \mathbb{R}^{|A|+|O|}$.  
\begin{align}
z_{\text{pair}} &= MLP_{\text{pair}}(z_2^D)\\
\tilde{z}_{\text{pair}} &= z_{\text{pair}} + \text{concat}(\tilde{y}_{\text{attr}},\tilde{y}_{\text{obj}})
\end{align}
To compute the composition pair predictions, $\tilde{z}_{\text{pair}}$ is fed to a Sparse Linear Layer. As illustrated in Figure \ref{fig:network}, pair output is calculated by a learnable weighted ($W^a\in \mathbb{R}^{|A|\cdot |O|},W^o\in \mathbb{R}^{|O|\cdot |A|}$) addition of corresponding attribute and object of each composition resulting in the compositional pair prediction $\tilde{y}_{\text{compose}} \in \mathbb{R}^{|A|\cdot |O|}$.
\small
\begin{align}
\tilde{y}_{\text{compose}}[a_i,o_j]&= \tilde{z}_{\text{pair}}[a_i]\odot W_{a_i,o_j}^a+\tilde{z}_{\text{pair}}[o_j]\odot W_{o_j,a_i}^o
\end{align}
\normalsize
Where, $W^a\in \mathbb{R}^{|A|\cdot |O|}, W^o\in \mathbb{R}^{|O|\cdot |A|}$, $i\in(1,...,|A|)$ and $j\in(1,...,|O|)$.
The proposed Sparse Linear Layer requires only $2(|A|\cdot|O|)$ number of learnable parameters while a standard linear layer would require $(|A|+|O|)(|A|\cdot|O|)$ number of parameters.
Lastly, we combine both the decomposition pair prediction and compositional pair prediction to produce the final pair prediction
\begin{align}
\tilde{y} &= \tilde{y}_{\text{decompose}}+\eta \tilde{y}_{\text{compose}}
\end{align}
where $\eta$ is the scale factor.

\subsection{Classification Head}
For classification of attributes, objects and pairs, we extract the three $\mathtt{[class]}$ tokens from the last layer output $z^D$ of the transformer network and process each token through SLC to get the final  pair prediction $\tilde{y} \in \mathbb{R}^{(|A|\cdot |O|)}$, attributes prediction $\tilde{y}_{\text{attr}} \in \mathbb{R}^{|A|}$ and objects predictions $\tilde{y}_{\text{obj}} \in \mathbb{R}^{|O|}$ as auxiliary outputs.  
\begin{align}
\tilde{y},\tilde{y}_{\text{attr}},\tilde{y}_{\text{obj}} &=SLC(z_0^D,z_1^D,z_2^D)
\end{align}
In the context of OW-CZSL, label space expands proportionally to $|A|\cdot|O|$, potentially resulting in a substantially large output space. In order to mitigate this computational overhead, we deploy similar filtering schema as previous OW-CZSL works\cite{karthik2022kg,nayak2022learning} to filter-out unfeasible compositions. We aggregate text embedding for each word in the vocabulary by utilizing ConceptNet numberbatch \cite{speer2017conceptnet} and GloVe \cite{pennington2014glove} embeddings. Subsequently, for each composition a feasibility score is calculated by taking the average of cosine similarity between respective attribute and object. For each pair, the maximum of two similarity scores is set as the final feasibility score. We set the threshold for the model feasibility by empirically adjusting the value.
Using the threshold values, we create a binary mask, $f_{pair} \in \mathbb{R}^{(|A|\cdot |O|)}$ which performs element-wise multiplication with the output from SLC $\tilde{y} \in \mathbb{R}^{(|A|\cdot |O|)}$ to produce the final compositional output $\bar{y} \in \mathbb{R}^{(|A|\cdot |O|)}$.
\begin{align}
\bar{y} &= \tilde{y} \cdot f_{pair}
\end{align}
\subsection{Training Objectives}
\bgroup
\def\arraystretch{1.2}
\begin{table}[t]
\setlength\tabcolsep{2.5pt}
\fontsize{7}{9}\selectfont 
\caption{Summary of dataset statistics where seen and unseen compositions are denoted by Y$^s$ and Y$^u$}
\begin{center}
\label{tab:dataset}
\begin{tabular}{l c c c ccc c c c}
\cline{2-9}
 &  & Train set&& && Val set && Test set \\
\cline{2-5}\cline{7-7}\cline{9-9}
Datasets& attr. & obj.& attr.$\times$ obj. & Y$^s$  && Y$^s$/Y$^u$ && Y$^s$/Y$^u$ \\
\cline{1-9}
MIT-states\cite{isola2015discovering} & 115 & 245 & 28175 & 1262 &&300 / 300&  & 400 / 400 \\
C-GQA\cite{mancini2021open} & 413 & 674 & 278362 & 5592 &  & 1252 / 1040&& 888 / 923 \\
VAW-CZSL\cite{Saini2022Disentangling} & 440 & 541 & 238040 & 11175 &  &2121 / 2322& & 2449 / 2470\\
\cline{1-9}
\end{tabular}
\end{center}
\end{table}
\egroup
Compared to conventional OW-CZSL setting, rather than computing cosine similarity in an embedding space, the proposed method utilizes cross-entropy loss over predictions.

\noindent\textbf{Pair Loss:} We compute the pair loss by calculating cross entropy loss over the final pair predictions $\bar{y}$ with pair ground-truth $y$.

\noindent\textbf{TopK embedding Loss:} We compute cross-entropy loss between the highest attribute and objects attention scores obtained from the TopK selection module and the attribute and object ground-truths $y_{attr}$ and $y_{obj}$.

\noindent\textbf{Disentangling Loss:} 
We empirically observe that by utilizing separate attribute and object cross-entropy losses for $\tilde{y}_{attr}$ and $\tilde{y}_{obj}$ with the ground-truths $y_{attr}$ and $y_{obj}$ enhances the proposed method's ability in achieving disentanglement.

A combined loss function $\mathcal {L}$ is minimized over all the training images to train the proposed method end-to-end manner. The weights for each loss ($\alpha_i,i=1,2,3$) are empirically computed.
 \begin {align} 
 \mathcal {L} &= \mathcal {L}_{Pair} + \alpha _1 \mathcal {L}_{TopK} + \alpha _2 \mathcal {L}_{Attr} + \alpha _3 \mathcal {L}_{Obj} 
 \end {align}

\section{Experiment}
\label{sec:experiment}
\newcolumntype{K}{>{\centering\arraybackslash}p{0.7cm}}
\def\arraystretch{1.3}
\begin{table*}[t]
\setlength\tabcolsep{1.2pt}
\caption{Open world performance on MIT-States, C-GQA and VAW-CZSL. As evaluation matrices we refer to AUC with seen and unseen compositions with different bias terms along with HM.}
\begin{center}
\fontsize{8}{9}\selectfont 
\label{tab:results1}
\begin{tabular}{l c K K K K|K K K K|K K K K}
Method &Backbone & \multicolumn{4}{c|}{MIT} &  \multicolumn{4}{c|}{C-GQA} & \multicolumn{4}{c}{VAW-CZSL} \\
&& S & U  & HM  & AUC & S & U  & HM  & AUC& S & U  & HM  & AUC\\
\cline{1-14}
TMN \cite{purushwalkam2019task}&R18& 12.6& 0.9& 1.2& 0.1& - & - & - & - & - & - & - & -\\
VisProd \cite{misra2017red}&R18&20.9& 5.8& 5.6& 0.7& 24.8& 1.7& 2.8& 0.33& -& -& -& - \\
SymNet\cite{li2020symmetry}&R18 &21.4& 7.0 &5.8& 0.8& 26.7 &2.2& 3.3& 0.43& -& -& -& - \\
ComCos\cite{mancini2021open}&R18 & 25.4 & 10.0 & 8.9 & 1.6 &  28.4& 1.8 & 2.8 & 0.39&4.3 & 1.0& 1.1& 0.03\\
CGE\cite{naeem2021learning}&R18 & 32.4 & 5.1 & 6.0 & 1.0 &  32.7& 1.8 & 2.9 & 0.47&8.6 &2.8&2.2 & 0.13\\
KG-SP\cite{karthik2022kg}&R18 & 28.4 & 7.5 & 6.7 & 1.3  &  31.5 & 2.9 &4.7  &0.78&6.4 &2.4 &1.8 &0.08 \\
SAD-SP\footnotemark\addtocounter{footnote}{-1}\cite{li2023distilled}&R18& 29.1 &7.6 &7.8& 1.4 &31.0&3.9&5.9&1.0& -& -& -& -\\
DRANet\footnotemark\cite{liu2023simple}&R18& 29.8 &7.8 &7.9& 1.5  &31.3&3.9&6.0&1.05& -& -& -& - \\
\cdashline{1-14}   
ADE\cite{hao2023learning}&ViT-B & -& -& -&-&35.1&4.8&7.6&1.42& -& -& -&-\\
KG-SP$_{vit}$&ViT-B & 28.6  & 11.8 & 10.3 & 2.1 & 31.3 & 3.4 & 5.1 &0.87&15.0&4.5 &3.9&0.38 \\
\cline{1-14}
Ours&ViT-B& \textbf{36.3} & \textbf{12.5} & \textbf{12.4} & \textbf{3.1}  & \textbf{35.8}& \textbf{5.6} & \textbf{7.8} &\textbf{1.6} & \textbf{16.5} & \textbf{6.7}& \textbf{6.7} & \textbf{0.82}\\
\cline{1-14}
\end{tabular}
\end{center}
\end{table*}
\subsection{Datasets and Metrics}
When evaluating the proposed model, we refer to three datasets namely, MIT-states\cite{isola2015discovering},  C-GQA\cite{mancini2021open} and newly introduced VAW-CZSL \cite{Saini2022Disentangling}. MIT-states contains 53753 web images with common objects (e.g., balloon) and their states (e.g., filled). C-GQA is created based
on the Stanford GQA dataset \cite{hudson2019gqa}, in which C-GQA consists of 39298 images with a higher number of attributes and everyday objects (e.g., bowl) compared to previous datasets. Lastly, VAW-CZSL with 92583 images which is a sub-set of VAW\cite{pham2021learning} is introduced to mitigate the issues of having a low number of image samples. A summary of each dataset is shown in Table \ref{tab:dataset}. Moreover, we used the train:val:test splits proposed by Purushwalkam \et \cite{purushwalkam2019task} and Saini \et \cite{Saini2022Disentangling}.
\vspace{-6mm}
\subsubsection{Evaluation} To evaluate the proposed method, we use the general OW-CZSL setup suggested by Purushwalkam \et \cite{purushwalkam2019task} to combine with the dataset statistics. In order to alleviate the bias against unseen compositions, we follow the same protocol to add a scalar bias term to calibrate the model as the scalar bias varies from negative infinity to positive infinity. We present the best seen accuracy (S), best unseen accuracy (U) and the Area Under the Curve (AUC). Lastly, the Harmonic mean (HM) is reported to balance out the bias.
\vspace{-4mm}
\subsubsection{Implementation}
\vspace{-2mm}
For all experiments, our model is trained with the AdamW optimizer\cite{loshchilov2018fixing} with a base learning rate of 3.5$e^{-5}$ with layer-wise learning rate decay\cite{zhang2020revisiting} and weight decay of 10$e^{-2}$. The learning rate is warmed up with 20\% of the total training steps and decayed in a 0.5$\times$cosine cycle. We follow similar data augmentations as CoT\cite{kim2023hierarchical}, such that input images are augmented with random crop and horizontal flip, and resized into 224$\times$224. Due to the dual modality nature of the experiments, we fine-tune all the layers of the ViT encoder. We fine-tune the proposed model for 20 epochs for VAW-CZSL and 10 epochs for other three datasets on four NVIDIA RTX A4000 GPUs with a batch size of 64. Empirical analysis on learning rates, weights for each loss $\alpha_i$ and the effect of number of frozen layers in ViT for each dataset can be found in the supplementary.
\footnotetext{No code base provided to reproduce the results.}
\subsection{Results}
Table \ref{tab:results1} presents a comprehensive summary of the primary experiments and compares the performance of the proposed model against that of various baseline models followed by an ablation study on design choices and qualitative analysis of the results.
\newcommand{\green}[1]{{\tiny\textcolor{green}{#1}}}
\newcommand{\red}[1]{{\tiny\textcolor{red}{#1}}}
\def\arraystretch{1.2}
\begin{table*}[t]
\setlength\tabcolsep{3pt}
\fontsize{8}{9}\selectfont 
\caption{Ablation studies results. Numbers in italics refer to individual ablation experiments. \textit{(1)} denotes our baseline model.}
\begin{center}
\label{tab:results_ab}
\begin{tabular}{l c c c c}
Variation& S & U  & HM  & AUC \\
\cline{1-5}
\textit{(1)} Baseline \textit{(K)}=3&  36.3 & 12.5 & 12.4 & 3.1 \\
\textit{(2)} Removing TopK Selection Module& 32.2\red{(-4.1)} & 10.4\red{(-2.1)} & 10.1\red{(-2.3)} & 2.2\red{(-0.9)} \\
\textit{(3)} TopK Selection Module \textit{(K)}=1& 36.0\red{(-0.3)} & 12.3\red{(-0.2)} & 12.2\red{(-0.2)} & 3.0\red{(-0.1)} \\
\textit{(4)} TopK Selection Module \textit{(K)}=5& 35.5\red{(-0.8)} & 12.1\red{(-0.4)} & 12.0\red{(-0.4)} & 2.9\red{(-0.2)} \\
\textit{(5)} TopK Selection Module \textit{(K)}=10& 34.9\red{(-1.4)} & 12.3\red{(-0.2)} & 12.1\red{(-0.3)} & 2.9\red{(-0.2)} \\
\cline{1-5}
\end{tabular}
\end{center}
\vspace{-1mm}
\def\arraystretch{1.2}
\setlength\tabcolsep{1.2pt}
\caption{Performance comparison between SLC and conventional FC on MIT-States, C-GQA and VAW-CZSL.}
\begin{center}
\fontsize{7}{9}\selectfont 
\label{tab:slc}
\begin{tabular}{l c c c c c|cc c c c|cc c c c}
Method  & \multicolumn{5}{c|}{MIT} &  \multicolumn{5}{c|}{C-GQA} & \multicolumn{5}{c}{VAW-CZSL} \\
&\# of params& S & U  & HM  & AUC &\# of params& S & U  & HM  & AUC&\# of params& S & U  & HM  & AUC\\
\cline{1-16}
With SLC &0.6M&  {36.3} &  {12.5} &  {12.4} &  {3.1}  &2.5M&  {35.8}&  {5.6} &  {7.8} & {1.6} &2.2M&  {16.5} &  {6.7}&  {6.7} &  {0.82}\\
With FC&10.7M&  {35.5} &  {10.7} &  {11.1} &  {2.6}  &304.5M&  {35.9}&  {4.2} &  {6.1} & {1.2} &235.2M&  {12.7} &  {2.0}&  {2.8} &  {0.2}\\
\cline{1-16}
\end{tabular}
\end{center}
\vspace{-1mm}
\setlength\tabcolsep{3pt}
\caption{Comparison between Proposed model and CLIP. As evaluation matrices we refer to AUC with seen and unseen accuracies and HM.}
\begin{center}
\fontsize{7}{9}\selectfont 
\label{tab:clip}
\begin{tabular}{l llc c c c|c c c c|c c c c}
Method &Total &\# Train & \multicolumn{4}{c|}{MIT-States}&\multicolumn{4}{c|}{C-GQA}& \multicolumn{4}{c}{VAW-CZSL} \\
&\# Par.&Samples& S & U  & HM  & AUC& S & U  & HM  & AUC & S & U  & HM  & AUC\\
\cline{1-15}
CLIP\cite{radford2021learning}&428M&13B & 30.1 & 14.3 & 12.8 & 3.0 & 7.5& 4.6 & 4.0 &0.27&4.2 &\textbf{9.8} & 3.9&0.28\\
$\mathbb{CSP}$\cite{nayak2022learning}& 428M&13B&\textbf{46.3} & \textbf{15.7} & \textbf{17.4} & \textbf{5.7} &28.7& 5.2 & 6.9 &1.2& 4.0& 9.4 & 3.6 & 0.26\\
\cline{1-15}
Ours&102M&14M&  36.3 & 12.5 & 12.4 & 3.1&\textbf{35.8}& \textbf{5.6} & \textbf{7.8} &\textbf{1.6}& \textbf{16.5} & 6.7& \textbf{6.7} & \textbf{0.82}\\
\cline{1-15}
\end{tabular}
\end{center}
\end{table*}
\vspace{-1mm}
\subsubsection{Baselines}
We compare the proposed method with networks derived from CZSL \cite{naeem2021learning,misra2017red,purushwalkam2019task} and networks designed for OW-CZSL \cite{karthik2022kg,liu2023simple,li2023distilled}. Previous baseline models consist of ResNet-18\cite{he2016deep} based image encoders. As a fair assessment of our model performance, we substitute the ResNet-18 encoder in KS-SP \cite{karthik2022kg} with a ViT-B/16 encoder\footnote{We trained the model for 200 numbers of epochs with provided hyper parameters.}. Other baseline results were obtained from \cite{karthik2022kg,liu2023simple,li2023distilled} or re-produced from an official code base. Moreover, we compare the proposed method with CLIP \cite{radford2021learning} based models in Section \ref{sec:clip}.
\vspace{-0.7cm}
\subsubsection{Results on MIT-States}
\vspace{-0.2cm}
Even though, MIT-states contains label noise\cite{atzmon2020causal}, (Qualitative analysis can be found in the supplementary.) considers as a foundational dataset for OW-CZSL. Our model shows better performance over seen accuracy with a 7.7\% margin along with a 2.1\% increment in HM and a 1\% increment in AUC demonstrating higher performance compared to KG-SP$_{vit}$. Nevertheless, the proposed model shows significant improvements over other baseline models as well.
\vspace{-0.3cm}
\subsubsection{Results on C-GQA}
\vspace{-0.2cm}
Amidst C-GQA containing a significantly higher number of compositions, the proposed model was able to achieve state-of-the-art performance with a 4.1\% seen accuracy improvement in addition to 2.2\% increment in HM over KG-SP$_{vit}$, a resulting gain in AUC of 0.6\%.
\vspace{-0.3cm}
\subsubsection{Results on VAW-CZSL}
\vspace{-0.2cm}
As an alternative, more refined dataset compared to C-GQA, we assess the performance of our model on VAW-CZSL.\cite{Saini2022Disentangling}. The proposed model was able to achieve 9.5\% increment on seen accuracy while maintaining on par unseen accuracy with KG-SP$_{vit}$, a substantial improvement of 0.6\% in AUC while gaining 3.5\% in HM.
\subsection{ Ablation Studies}
Purpose model incorporates several design choices and we present empirical evidence to substantiate the aforementioned choices and alternatives. We discuss each experiment in detail below.
\vspace{-0.35cm}
\subsubsection{Effect of TopK Selection Module}
\vspace{-0.1cm}
One of the pivotal components of the proposed method lie in the TopK selection module, which leverages semantic information provided by text data. To evaluate its impact, we omit this module from the network architecture and train the resulting network using similar hyper-parameters. All the experiments are conducted on MIT-states\cite{isola2015discovering}, with fixed random seed initialization. \textit{(2)} illustrates that the TopK selection module provides a significant improvement over all four evaluation matrices.

Subsequently, we follow up with the results obtained from \textit{(2)} by varying the number of selected text embeddings\textit{(K)}. As shown in \textit{(3)}, when constraining the module to exclusively select one attribute and one object, if correct options are retained in secondary scores, it induces a preemptive failure within the model.
However as indicated in \textit{(4)} and \textit{(5)}, introducing an excessive number of attribute and object selections may overwhelm the input with choices thereby resulting in a degradation of accuracy. This serves as an indicator that beyond a certain threshold of selectors, it may no longer be beneficial to retain the textual knowledge. 
Consequently, this prompts a utilization of the TopK selection module not as a classifier, but rather as a selection module to identify suitable candidates for the transformer encoder. Empirically, we choose \textit{K=3} as the optimal number of attributes and object candidates for the proposed model.
\vspace{-0.1cm}
\def\boxit#1{%
  \smash{\color{black}\fboxrule=0.5pt\relax\fboxsep=1.8pt\relax%
  \llap{\rlap{\fbox{\vphantom{0}\makebox[#1]{}}}~}}\ignorespaces
}
\newcommand{\correct}[1]{{\textcolor{blue}{#1}}}
\newcommand{\wrong}[1]{{\textcolor{red}{#1}}}
\newcommand{\okay}[1]{{\textcolor{purple}{#1}}}
\newcolumntype{C}[1]{>{\centering\arraybackslash}m{#1}}
\newcommand*\rot{\rotatebox[origin=c]{90}}
\begin{table*}[h]
\setlength\extrarowheight{2pt} 
\centering
\tiny
\caption{Qualitative results for proposed method. First column: correct pair prediction; Second column: correct attribute prediction; Third column: correct object prediction; Fourth column: incorrect prediction. Visually sound predictions are highlighted in \wrong{red}}\label{fig:quality}
\begin{tabular}{| C{0.05cm} | C{0.4cm} C{0.4cm} C{1.2cm} | C{0.5cm} C{0.5cm} C{1.3cm}| C{0.4cm} C{0.5cm} C{1.3cm}|C{0.5cm} C{0.4cm} C{1.4cm}|}
  \cline{2-13}
  \multicolumn{1}{c|}{} 
  & \multicolumn{3}{c|}{{Correct Pair}} & \multicolumn{3}{c|}{{Correct Attribute}} & \multicolumn{3}{c|}{{Correct Object}}& \multicolumn{3}{c|}{{Incorrect Pair}} \\[0.5ex] 
  \cline{1-13}  
  \multirow{3}{*}[-6ex]{\rot{MIT-States}}
  & \multicolumn{3}{c|}{GT: Mossy Pond}
  &\multicolumn{3}{c|}{GT: Small Building}
  &\multicolumn{3}{c|}{GT: Curved Road}
  &\multicolumn{3}{c|}{GT: Grimy Shower}\\
  
  &\multicolumn{3}{c|}{\centering\includegraphics[width=0.13\linewidth, height= 1.3cm]{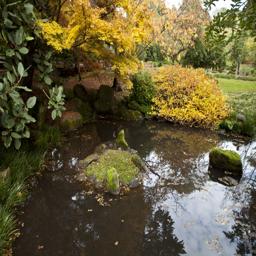}}
  &\multicolumn{3}{c|}{\centering\includegraphics[width=0.13\linewidth, height= 1.3cm]{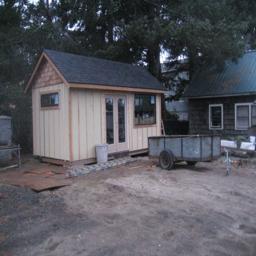}}
  &\multicolumn{3}{c|}{\centering\includegraphics[width=0.13\linewidth, height= 1.3cm]{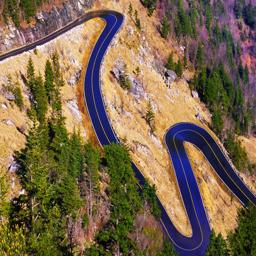}}
  &\multicolumn{3}{c|}{\centering\includegraphics[width=0.13\linewidth, height= 1.3cm]{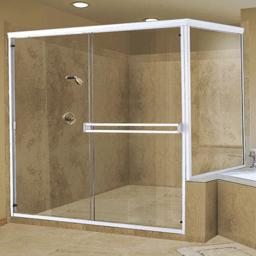}}\\
     
  &\boxit{1in}\scalebox{.8}{\text{\correct{mossy}}}&\scalebox{.8}{\text{\correct{pond}}}&\scalebox{.8}{\text{\correct{mossy pond}}}
  &\boxit{1.05in}\scalebox{.8}{\text{\correct{small}}}&\scalebox{.8}{\text{\wrong{house}}}&\scalebox{.8}{\text{\wrong{small house}}}
  &\boxit{1.05in}\scalebox{.8}{\text{\wrong{barren}}}&\scalebox{.8}{\text{\correct{road}}}&\scalebox{.8}{\text{\wrong{barren road}}}
  &\boxit{1.1in}\scalebox{.8}{\text{small}}&\scalebox{.8}{\text{\wrong{bathroom}}}&\scalebox{.8}{\text{small bathroom}}\\
\addlinespace[-1ex]
  &\scalebox{.8}{\text{large}}&\scalebox{.8}{\text{lake}}&\scalebox{.8}{\text{large pond}}
  &\scalebox{.8}{\text{large}}&\scalebox{.8}{\text{\correct{building}}}&\scalebox{.8}{\text{\wrong{weathered house}}}
  &\scalebox{.8}{\text{\wrong{verdant}}}&\scalebox{.8}{\text{highway}}&\scalebox{.8}{\text{\wrong{verdant road}}}
  &\scalebox{.8}{\text{large}}&\scalebox{.8}{\text{\correct{shower}}}&\scalebox{.8}{\text{large bathroom}}\\
  \addlinespace[-1ex]
  &\scalebox{.8}{\text{\wrong{verdant}}}&\scalebox{.8}{\text{stream}}&\scalebox{.8}{\text{\wrong{small pond}}}
  &\scalebox{.8}{\text{\wrong{weathered}}}&\scalebox{.8}{\text{\wrong{farm}}}&\scalebox{.8}{\text{large house}}
  &\scalebox{.8}{\text{mossy}}&\scalebox{.8}{\text{\wrong{valley}}}&\scalebox{.8}{\text{barren highway}}
  &\scalebox{.8}{\text{\wrong{empty}}}&\scalebox{.8}{\text{cabinet}}&\scalebox{.8}{\text{\wrong{empty bathroom}}}\\ 
  \cline{1-13}
 \multirow{3}{*}[-8ex]{\rot{C-GQA}}
  & \multicolumn{3}{c|}{GT: Chocolate Cake}
  &\multicolumn{3}{c|}{GT: Brick Ground}
  &\multicolumn{3}{c|}{GT: Dark Cloud}
  &\multicolumn{3}{c|}{GT: Brown Dirt}\\
  
  &\multicolumn{3}{c|}{\centering\includegraphics[width=0.13\linewidth, height= 1.3cm]{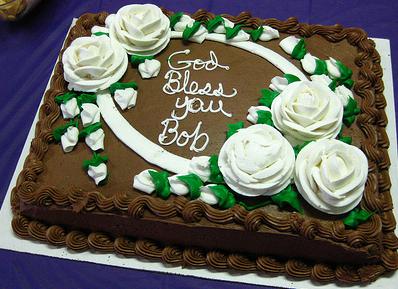}}
  &\multicolumn{3}{c|}{\centering\includegraphics[width=0.13\linewidth, height= 1.3cm]{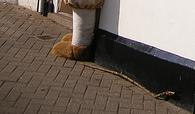}}
  &\multicolumn{3}{c|}{\centering\includegraphics[width=0.13\linewidth, height= 1.3cm]{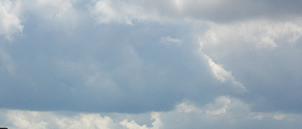}}
  &\multicolumn{3}{c|}{\centering\includegraphics[width=0.13\linewidth, height= 1.3cm]{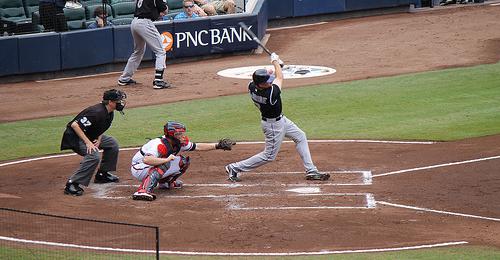}}\\
 
  &\boxit{1.035in}\scalebox{.8}{\text{\wrong{brown}}}&\scalebox{.8}{\text{\correct{cake}}}&\scalebox{.8}{\text{\correct{chocolate cake}}}
  &\boxit{1.05in}\scalebox{.8}{\text{\correct{brick}}}&\scalebox{.8}{\text{\wrong{sidewalk}}}&\scalebox{.8}{\text{\wrong{brick sidewalk}}}
  &\boxit{1in}\scalebox{.8}{\text{\wrong{large}}}&\scalebox{.8}{\text{\correct{cloud}}}&\scalebox{.8}{\text{\wrong{large cloud}}}
  &\boxit{1.12in}\scalebox{.8}{\text{\wrong{baseball}}}&\scalebox{.8}{\text{\wrong{game}}}&\scalebox{.8}{\text{\wrong{baseball game}}}\\
\addlinespace[-1ex]
  &\scalebox{.8}{\text{\correct{chocolate}}}&\scalebox{.8}{\text{\wrong{frosting}}}&\scalebox{.8}{\text{\wrong{brown cake}}}
  &\scalebox{.8}{\text{\wrong{gray}}}&\scalebox{.8}{\text{\correct{ground}}}&\scalebox{.8}{\text{\correct{brick ground}}}
  &\scalebox{.8}{\text{\wrong{gray}}}&\scalebox{.8}{\text{smoke}}&\scalebox{.8}{\text{\wrong{gray cloud}}}
  &\scalebox{.8}{\text{\correct{brown}}}&\scalebox{.8}{\text{\wrong{stadium}}}&\scalebox{.8}{\text{\wrong{baseball stadium}}}\\
\addlinespace[-1ex]
  &\scalebox{.8}{\text{\wrong{white}}}&\scalebox{.8}{\text{\wrong{table}}}&\scalebox{.8}{\text{white cake}}
  &\scalebox{.8}{\text{concrete}}&\scalebox{.8}{\text{\wrong{walkway}}}&\scalebox{.8}{\text{\wrong{brick walkway}}}
  &\scalebox{.8}{\text{\correct{dark}}}&\scalebox{.8}{\text{mountain}}&\scalebox{.8}{\text{\correct{dark cloud}}}
  &\scalebox{.8}{\text{\wrong{squatting}}}&\scalebox{.8}{\text{\correct{dirt}}}&\scalebox{.8}{\text{baseball dirt}}\\ 
  \cline{1-13}
  
  \multirow{3}{*}[-5ex]{\rot{VAW-CZSL}}
  & \multicolumn{3}{c|}{GT: Ripe Apple}
  &\multicolumn{3}{c|}{GT: Red Box}
  &\multicolumn{3}{c|}{GT: In the Air Kite}
  &\multicolumn{3}{c|}{GT: Standing Zebra}\\
  
  &\multicolumn{3}{c|}{\centering\includegraphics[width=0.13\linewidth, height= 1.3cm]{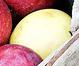}}
  &\multicolumn{3}{c|}{\centering\includegraphics[width=0.13\linewidth, height= 1.3cm]{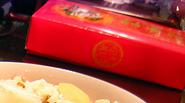}}
  &\multicolumn{3}{c|}{\centering\includegraphics[width=0.13\linewidth, height= 1.3cm]{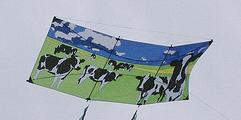}}
  &\multicolumn{3}{c|}{\centering\includegraphics[width=0.13\linewidth, height= 1.3cm]{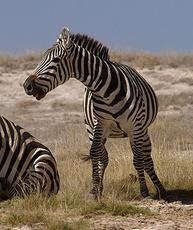}}\\

  &\boxit{1in}\scalebox{.8}{\text{\correct{ripe}}}&\scalebox{.8}{\text{\correct{apple}}}&\scalebox{.8}{\text{\correct{ripe apple}}}
  &\boxit{1in}\boxit{1in}\scalebox{.8}{\text{\correct{red}}}&\scalebox{.8}{\text{\wrong{table}}}&\scalebox{.8}{\text{red tray}}
  &\boxit{1in}\scalebox{.8}{\text{\wrong{green}}}&\scalebox{.8}{\text{\correct{kite}}}&\scalebox{.8}{\text{\wrong{green kite}}}
  &\boxit{1in}\scalebox{.8}{\text{\wrong{black}}}&\scalebox{.8}{\text{\wrong{grass}}}&\scalebox{.8}{\text{\wrong{dry grass}}}\\
  \addlinespace[-1ex]
  &\scalebox{.8}{\text{\wrong{red}}}&\scalebox{.8}{\text{\wrong{wood}}}&\scalebox{.8}{\text{full apple}}
  &\scalebox{.8}{\text{\wrong{pink}}}&\scalebox{.8}{\text{\wrong{plate}}}&\scalebox{.8}{\text{red bowl}}
  &\scalebox{.8}{\text{\wrong{black}}}&\scalebox{.8}{\text{flag}}&\scalebox{.8}{\text{\wrong{blue kite}}}
  &\scalebox{.8}{\text{\wrong{white}}}&\scalebox{.8}{\text{\correct{zebra}}}&\scalebox{.8}{\text{\wrong{tall grass}}}\\
 \addlinespace[-1ex]
  &\scalebox{.8}{\text{\wrong{full}}}&\scalebox{.8}{\text{orange}}&\scalebox{.8}{\text{\wrong{round apple}}}
  &\scalebox{.8}{\text{silver}}&\scalebox{.8}{\text{tray}}&\scalebox{.8}{\text{red table}}
  &\scalebox{.8}{\text{\wrong{blue}}}&\scalebox{.8}{\text{\wrong{string}}}&\scalebox{.8}{\text{\wrong{flying kite}}}
  &\scalebox{.8}{\text{\correct{standing}}}&\scalebox{.8}{\text{\wrong{stripe}}}&\scalebox{.8}{\text{white grass}}\\ 
\cline{1-13}
\end{tabular}
\end{table*}
\vspace{-0.1cm}
\subsubsection{Effect of Sparse Linear Compositor}
\vspace{-0.1cm}
To demonstrate the efficiency of the proposed SLC in comparison to a conventional FC layer, we conduct an ablation study where we replace the Sparse Linear layer with a FC layer that connects concatenated attribute and object prediction vectors to pair output predictions in a fully connected manner. 
Table \ref{tab:slc} demonstrates remarkable performance of SLC while requiring only a fraction of the parameters necessary for a FC layer. 
From experiments with C-GQA and VAW-CZSL, it becomes evident that as the count of attributes and objects increases, there is a corresponding increment in the required number of full connections, characterized by $\mathcal{O}((|A|+|O|)(|A|\cdot|O|))$.
Notably, the proposed model achieves 1.7\% increment in HM for C-GQA while utilizing only 0.8\% of the parameters compared to a FC layer which can be contributed to overfitting that occurred during training.



\subsection{Qualitative results}
\label{sec:qual}
Figure \ref{fig:quality} illustrate qualitative results of the proposed method. For each dataset, we present four samples. Column 1 displays the correct composition predictions while columns 2 and 3 show accurate attribute and object predictions. Furthermore, final column consists of incorrect predictions. For each cell, three columns illustrate top-3 results for attributes, objects, and compositions, respectively, with ground-truth label positioned at the top (GT). We denote the final predictions using black boxes and predictions matching the ground truth are highlighted in blue. Moreover, predictions exhibiting visual correspondence with the image sample are highlighted in red. 
In all instances, the majority of the top-3 predictions describe visual content of the images acceptably well, albeit, in the last three cases, ground-truth label does not align with the final composition prediction. This prompts us to investigate predicting and evaluating the effect of an input image containing multiple objects as well as objects comprising multiple attributes. 
\subsection{Comparison with CLIP based Architectures}
\label{sec:clip}
To evaluate the effectiveness of the proposed model, we conduct experiments against CLIP \cite{radford2021learning} and CLIP based $\mathbb{CSP}$\cite{nayak2022learning} trained under the OW-CZSL setting. Experiments are performed across MIT-States and the two challenging datasets, C-GQA and VAW-CZSL. 
As shown in Table \ref{tab:clip}, proposed model outperform CLIP and $\mathbb{CSP}$ with a significant margin in C-GQA gaining 5\% seen accuracy and 1.2\% in HM compared to $\mathbb{CSP}$. For VAW-CZSL, increments in seen accuracy and HM are 12.8\% and 2.8\% respectively and our model was able to achieve competitive results on unseen accuracy compared to $\mathbb{CSP}$ and CLIP. While the proposed model achieves under par results compared to $\mathbb{CSP}$ on MIT-States, it surpasses CLIP with a significant margin. This demonstrates the robustness and scalability of the proposed model, indicating its adaptability in accommodating an increasing number of compositions.
Moreover, we highlight that we were able to achieve superior performance using only 25\% of parameters and 10\% of pre-training data compared to CLIP models. 



\vspace{-0.2cm}
\subsubsection{Limitations}
\label{sec:limit}
\vspace{-0.1cm}
Despite the proposed model yielding superior results, it struggles to accurately identify all the attributes associated with the given object thereby resulting in incorrect predictions as shown in Figure \ref{fig:quality} and in the Supplementary. Moreover, in-the-wild images often contain multiple object instances; therefore, it becomes necessary to extend the proposed model's capabilities to effectively address multiple objects while identifying multiple attributes of each object. 
\section{Conclusion and Future Work}
\label{sec:conclusion}
In this work, we propose a unified framework for OW-CZSL to increase the inter-modality interactions compared to shallow interactions present in prior works. The TopK selection module introduces the notion of text matching selection to CZSL research which alleviates the ambiguity present at the inference. Moreover, a sparse linear compositor allows the model to decompose and compose attributes and objects to leverage the generalizability. Through wide-ranging experiments, we demonstrate the capacity of our model while surpassing previous methods across three benchmark and achieving competitive results with CLIP based models. Lastly, we highlight limitations including tackling multiple primitives and entities simultaneously and improving feasibility computation which we leave for future works. The negative societal impact of our model is similar to zero-shot methods and is discussed in the Supplementary.
\section*{Acknowledgment}
We thank Trevine Oorloff, Vatsal Agarwal, Jaian Cuttari for meaningful discussions and valuable assistance.
{\small
\bibliographystyle{ieee_fullname}
\bibliography{PaperForReview}
}


\end{document}